\documentclass[twoside,9pt]{article}
\usepackage{extsizes}
\usepackage[super,sort&compress,comma]{natbib} 
\usepackage[version=3]{mhchem}
\usepackage[left=1.5cm, right=1.5cm, top=1.785cm, bottom=2.0cm]{geometry}
\usepackage{balance}
\usepackage{mathptmx}
\usepackage{sectsty}
\usepackage{graphicx} 
\usepackage{lastpage}
\usepackage[format=plain,justification=justified,singlelinecheck=false,font={stretch=1.125,small,sf},labelfont=bf,labelsep=space]{caption}
\usepackage{float}
\usepackage{fancyhdr}
\usepackage{fnpos}
\usepackage[english]{babel}
\addto{\captionsenglish}{%
  
}
\usepackage{array}
\usepackage{droidsans}
\usepackage{charter}
\usepackage[T1]{fontenc}
\usepackage[usenames,dvipsnames]{xcolor}
\usepackage{setspace}
\usepackage[compact]{titlesec}
\usepackage{hyperref}
\usepackage{amsmath}
\usepackage{amsfonts}

\usepackage{lineno}

\usepackage{epstopdf}

\definecolor{cream}{RGB}{222,217,201}

\begin{document}

\pagestyle{fancy}
\thispagestyle{plain}
\fancypagestyle{plain}{
\renewcommand{\headrulewidth}{0pt}
}

\makeFNbottom
\makeatletter
\renewcommand\LARGE{\@setfontsize\LARGE{15pt}{17}}
\renewcommand\Large{\@setfontsize\Large{12pt}{14}}
\renewcommand\large{\@setfontsize\large{10pt}{12}}
\renewcommand\footnotesize{\@setfontsize\footnotesize{7pt}{10}}
\makeatother

\renewcommand{\thefootnote}{\fnsymbol{footnote}}
\renewcommand\footnoterule{\vspace*{1pt}%
\color{cream}\hrule width 3.5in height 0.4pt \color{black}\vspace*{5pt}} 
\setcounter{secnumdepth}{5}

\makeatletter 
\renewcommand\@biblabel[1]{#1}            
\renewcommand\@makefntext[1]%
{\noindent\makebox[0pt][r]{\@thefnmark\,}#1}
\makeatother 
\renewcommand{\figurename}{\small{Fig.}~}
\sectionfont{\sffamily\Large}
\subsectionfont{\normalsize}
\subsubsectionfont{\bf}
\setstretch{1.125} 
\setlength{\skip\footins}{0.8cm}
\setlength{\footnotesep}{0.25cm}
\setlength{\jot}{10pt}
\titlespacing*{\section}{0pt}{4pt}{4pt}
\titlespacing*{\subsection}{0pt}{15pt}{1pt}

\fancyfoot{}
\fancyfoot[LO,RE]{\vspace{-7.1pt}\includegraphics[height=9pt]{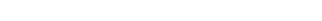}}
\fancyfoot[CO]{\vspace{-7.1pt}\hspace{13.2cm}\includegraphics{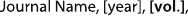}}
\fancyfoot[CE]{\vspace{-7.2pt}\hspace{-14.2cm}\includegraphics{head_foot/RF}}
\fancyfoot[RO]{\footnotesize{\sffamily{1--\pageref{LastPage} ~\textbar  \hspace{2pt}\thepage}}}
\fancyfoot[LE]{\footnotesize{\sffamily{\thepage~\textbar\hspace{3.45cm} 1--\pageref{LastPage}}}}
\fancyhead{}
\renewcommand{\headrulewidth}{0pt} 
\renewcommand{\footrulewidth}{0pt}
\setlength{\arrayrulewidth}{1pt}
\setlength{\columnsep}{6.5mm}
\setlength\bibsep{1pt}

\makeatletter 
\newlength{\figrulesep} 
\setlength{\figrulesep}{0.5\textfloatsep} 

\newcommand{\topfigrule}{\vspace*{-1pt}%
\noindent{\color{cream}\rule[-\figrulesep]{\columnwidth}{1.5pt}} }

\newcommand{\botfigrule}{\vspace*{-2pt}%
\noindent{\color{cream}\rule[\figrulesep]{\columnwidth}{1.5pt}} }

\newcommand{\dblfigrule}{\vspace*{-1pt}%
\noindent{\color{cream}\rule[-\figrulesep]{\textwidth}{1.5pt}} }

\newcommand{\yl}[1]{\textcolor{black}{#1}} 

\makeatother


\title{\textbf{TransPeakNet}: Solvent-Aware 2D NMR Prediction via Multi-Task Pre-Training and Unsupervised Learning$^\dag$} 

\author{Yunrui Li,$^{\ddag}$\textit{$^{a}$} Hao Xu,$^{\ddag}$\textit{$^{b}$} Ambrish Kumar,\textit{$^{c}$} Duosheng Wang,\textit{$^{d}$} Christian Heiss,\textit{$^{c}$} Parastoo Azadi,\textit{$^{c}$} and Pengyu Hong$^{\ast}$\textit{$^{a}$}}

\maketitle




\renewcommand*\rmdefault{bch}\normalfont\upshape
\rmfamily
\section*{}
\vspace{-1cm}


\footnotetext{\textit{$^{a}$~Department of Computer Science, Brandeis University, Waltham, MA, USA}\\
\textit{$^{b}$~Department of Medicine, Harvard Medical School, Boston, MA, USA}\\
\textit{$^{c}$~Complex Carbohydrate Research Center, University of Georgia, Athens, Georgia, USA}\\
\textit{$^{d}$~Department of Chemistry, Boston College, Chestnut Hill, MA, USA}}

\footnotetext{\ddag~These authors contributed equally to this work.}
\footnotetext{$^{\ast}$~Corresponding E-mail: hongpeng@brandeis.edu}

\begin{abstract}
    \textbf {Nuclear Magnetic Resonance (NMR) spectroscopy is essential for revealing molecular structure, electronic environment, and dynamics. Accurate NMR shift prediction allows researchers to validate structures by comparing predicted and observed shifts. While Machine Learning (ML) has improved one-dimensional (1D) NMR shift prediction, predicting 2D NMR remains challenging due to limited annotated data. To address this, we introduce an unsupervised training framework for predicting cross-peaks in 2D NMR, specifically Heteronuclear Single Quantum Coherence (HSQC).Our approach pretrains an ML model on an annotated 1D dataset of $^{1}\text{H}$ and $^{13}\text{C}$ shifts, then finetunes it in an unsupervised manner using unlabeled HSQC data, which simultaneously generates cross-peak annotations. Our model also adjusts for solvent effects. Evaluation on 479 expert-annotated HSQC spectra demonstrates our model's superiority over traditional methods (\textit{ChemDraw} and \textit{Mestrenova}), achieving Mean Absolute Errors (MAEs) of 2.05 ppm and 0.165 ppm for $^{13}\text{C}$ shifts and $^{1}\text{H}$ shifts respectively. Our algorithmic annotations show a 95.21\% concordance with experts' assignments, underscoring the approach's potential for structural elucidation in fields like organic chemistry, pharmaceuticals, and natural products.}
\end{abstract}

\section{Introduction}
Nuclear magnetic resonance (NMR) spectroscopy has emerged as a versatile tool with widespread applications across diverse scientific domains, including chemistry, environmental science, food science, material science, and drug discovery by unraveling molecular dynamics and structures.\cite{gunther1994nmr, claridge2016high, yu2021recent, lin2021high} The primary information of an NMR spectrum arises from the chemical shift, which is determined by the local environment of a nucleus and influenced by interactions through chemical bonds and space. This mechanism yields unique ``fingerprints'' corresponding to diverse functional groups or molecular motifs, thereby facilitating the streamlined deduction of atomic connectivity and arrangement. 

Interpreting NMR spectra requires following essential guidelines, often referred to as ``rules of thumb'', where specific chemical shifts are associated with distinctive functional groups.\cite{jonas2019rapid} The determination of molecular structures from varying chemical shifts on NMR spectra generally requires the expertise of experienced organic chemists. To facilitate the interpretation of NMR spectra, significant efforts have been directed towards computational simulation of NMR spectra.\cite{jonas2022prediction} Early computational approaches, like the Hierarchically Ordered Spherical Environment (HOSE) codes \cite{bremser1978hose}, aim to encapsulate atom neighborhoods in concentric spheres, utilizing a nearest-neighbor approach to predict NMR shift values. A recent HOSE approach \cite{kuhn2019stereo} yields Mean Absolute Errors (MAEs) of 3.52 ppm for $^{13}$C NMR and 0.29 ppm for $^{1}$H NMR on the nmrshiftdb2 \cite{steinbeck2003nmrshiftdb} dataset. Concurrently, significant efforts have been devoted to the ab initio calculation of NMR properties.\cite{lodewyk2012computational,willoughby2014guide} Density Functional Theory (DFT)-based methods were developed for certain small organic molecules, achieving MAEs of 2.9 ppm for $^{13}$C NMR and 0.23 ppm for $^{1}$H NMR.\cite{wiitala2006hybrid} However, the accuracy of these DFT-based methods relies heavily on the choice of the basis functions, which often require meticulous case-by-case manual tuning for each molecule. Moreover, the time-intensive nature of DFT calculations limits their applications to comprehensive and large datasets. Recently, the rise of Graph Neural Networks (GNN) and their successes in predicting molecular properties \cite{wieder2020compact, zhang2022graph, fang2022geometry, wang2023motif, xu2024graph} has prompted initiatives to employ GNNs for predicting peaks in NMR spectra \cite{jonas2019rapid, kwon2020neural, guan2021real}. The application of GNN to molecules is intuitive, as a molecular structure can be naturally represented as a graph, with each atom as a node and its chemical bonds as edges. On 1D NMR data, a GNN-based model achieves MAEs of 1.355 ppm for $^{13}$C NMR and 0.224 ppm for $^{1}$H NMR on the nmrshiftdb2 dataset.\cite{kwon2020neural} While considerable efforts have been made in developing predictive models for 1D NMR, the prediction of 2D NMR remains underexplored.

Heteronuclear Single Quantum Coherence (HSQC)  spectroscopy \cite{bodenhausen1980natural}, a sophisticated 2D NMR technique, is an important tool for elucidating atomic connectivity within complex molecules where conventional 1D NMR may prove insufficient.\cite{bross2005strategies,li2020practical} By correlating the chemical shifts of hydrogen nuclei with those of heteronuclear nuclei, typically carbon or nitrogen, via scalar coupling interactions, HSQC facilitates the comprehensive mapping of interatomic connections within a molecule. This mapping yields crucial insights into chemical bonding, molecular conformation, and intramolecular interactions. \textcolor{black}{As molecular structures become more complex, 1D NMR spectra tend to display increasingly overlapping peaks, making 2D NMR techniques such as HSQC essential for elucidating local structures.} A notable stride in this domain utilizes the ML approach to establish correlations between DFT-simulated HSQC spectra and empirical data to identify molecules \cite{zanardi2015giao}. However, the accurate prediction of HSQC spectra using ML techniques remains elusive, primarily due to the scarcity of large-scale, high-quality datasets, as well as the labor-intensive and time-consuming peak annotation process. While numerous annotated 1D spectra are available for training ML models, combining these results to reliably generate 2D NMR data is far from straightforward. As an example, a recent study introduced a method to integrate the state-of-the-art predictions of proton and carbon 1D spectra into HSQC spectra \cite{priessner2024}, achieving MAEs of 0.16 ppm for $^{1}\text{H}$ and 2.64 ppm for $^{13}\text{C}$. This study highlights the inherent difficulties in accurately predicting 2D NMR chemical shifts, as even though the selected 1D NMR models and methods achieve low error individually, such success cannot be transferred to HSQC cross-peak prediction. \textcolor{black}{Moreover, beyond the challenge of prediction, associating each HSQC peak with its corresponding local molecular structure is an inevitable task that requires expert domain knowledge and is often time-consuming. Proper calibration is typically necessary to ensure precise alignment between 1D and HSQC data before interpreting the HSQC spectra.} 

\begin{figure*}[!h]
\centering    
\includegraphics[width=0.98\linewidth]{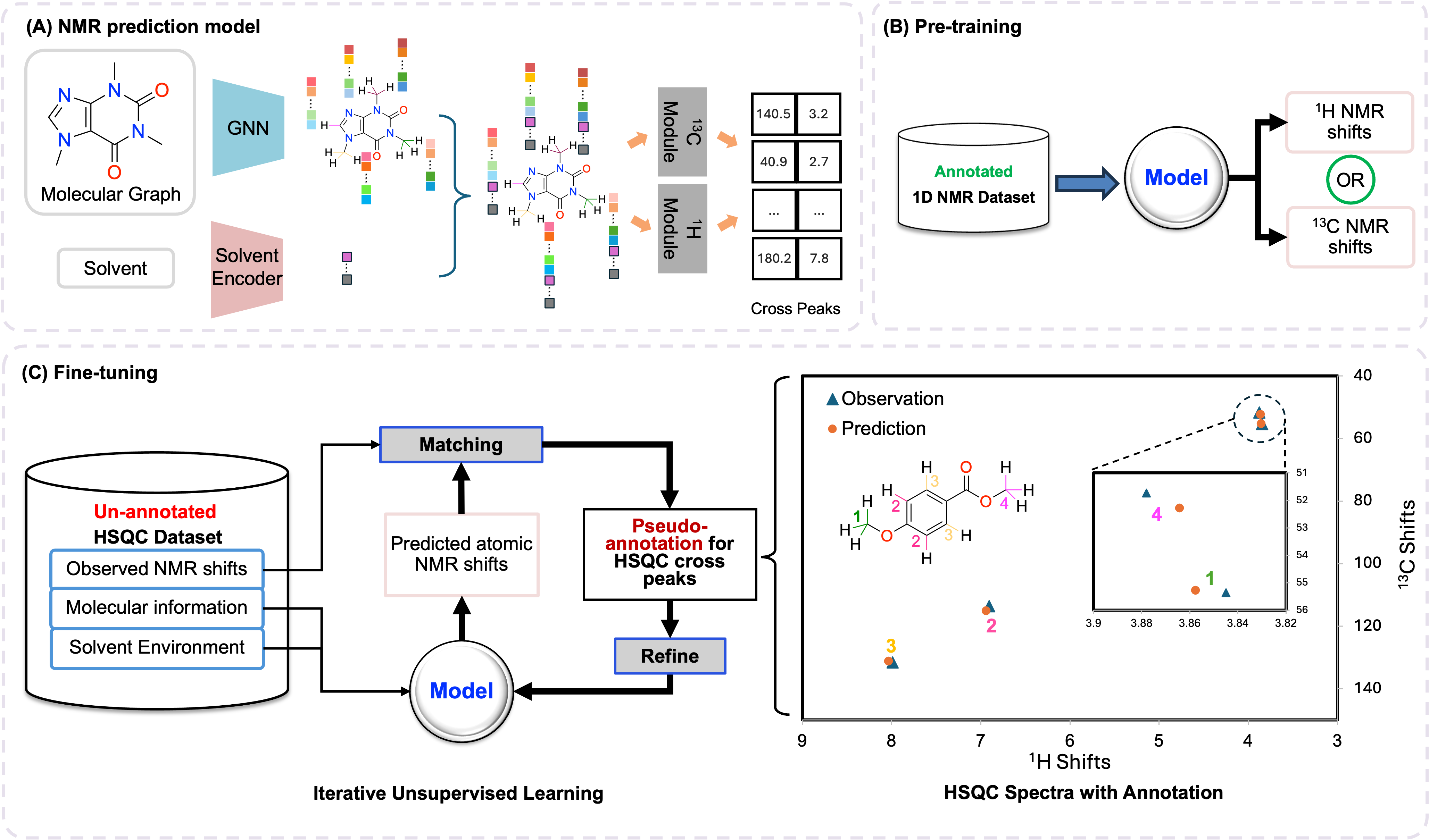}
\caption{Illustration of TransPeakNet model design (A) and training strategy ((B) and (C)). (A) The model takes a molecular structure and derives its atomic representations from a GNN. The solvent information is encoded into a latent representation via the Solvent encoder. The representation of each atom is concatenated with the solvent representation, which is then used to predict the cross shifts of carbon and proton. (B) Model pertaining on the annotated 1D NMR dataset using MTT. (C) The pre-trained model is refined through an unsupervised process using the unlabeled HSQC dataset. The final output of the model has both the HSQC cross-peaks and atom alignment.}
\label{fig:framework_training_strategy}
\end{figure*}

In light of the aforementioned challenges and opportunities in interpreting HSQC spectra, we propose \textbf{TransPeakNet}: a Transfer learning-based Peak prediction and assignment with unsupervised learning, illustrated in Figure \ref{fig:framework_training_strategy}. This framework enables end-to-end training and testing on experimental data. Once trained, the TransPeakNet model generates cross-peak predictions directly from a SMILES representation and the solvent environment used in the experiment. Additionally, it simultaneously associates each cross-peak signal with its corresponding carbon-proton pairs. Alongside a Graph Neural Network (GNN) module capturing structural nuances, the model incorporates a solvent encoder to effectively account for the impact of solvent environments on chemical shifts, which is essential for delivering accurate cross-peak prediction and peak assignment of HSQC spectra. To tackle the lack of annotated HSQC data, we designed a two-step transfer learning process. First, the model is pre-trained on a labeled 1D NMR dataset via Multi-Task pre-Training (MTT), enabling it to learn a wide range of C--H interactions. Then, we implement an unsupervised learning strategy that uses the unlabeled HSQC dataset to refine the model's ability to accurately discern and label HSQC cross peaks. 

The model is pre-trained using $\sim$24,000 annotated 1D NMR dataset from NMRShiftDB2\cite{steinbeck2003nmrshiftdb}, and finetuned on $\sim$19,000 experimental HSQC spectra from HMDB \cite{wishart2022hmdb} and CH-NMR-NP \cite{j-resonance}.The model is thoroughly evaluated and compared to traditional tools like \textit{ChemDraw}\cite{mills2006chemdraw} and \textit{Mestrenova}\cite{willcott2009mestre} using expert-annotated test datasets. On test dataset, the model achieves MAEs of 2.05 ppm and 0.165 ppm for $^{13}\text{C}$ shifts and $^{1}\text{H}$ shifts respectively. We also demonstrate that our model effectively considers the impact solvent has on chemical shift, when making the prediction. When compared with the traditional tools, our model shows promising improvements, especially as the molecular size becomes larger.

\section{Results}

\subsection{Performance on HSQC cross peak prediction and assignment}
\label{sc:general_performance}

\begin{figure}[!h]
    \centering
    \includegraphics[width=0.8\linewidth]{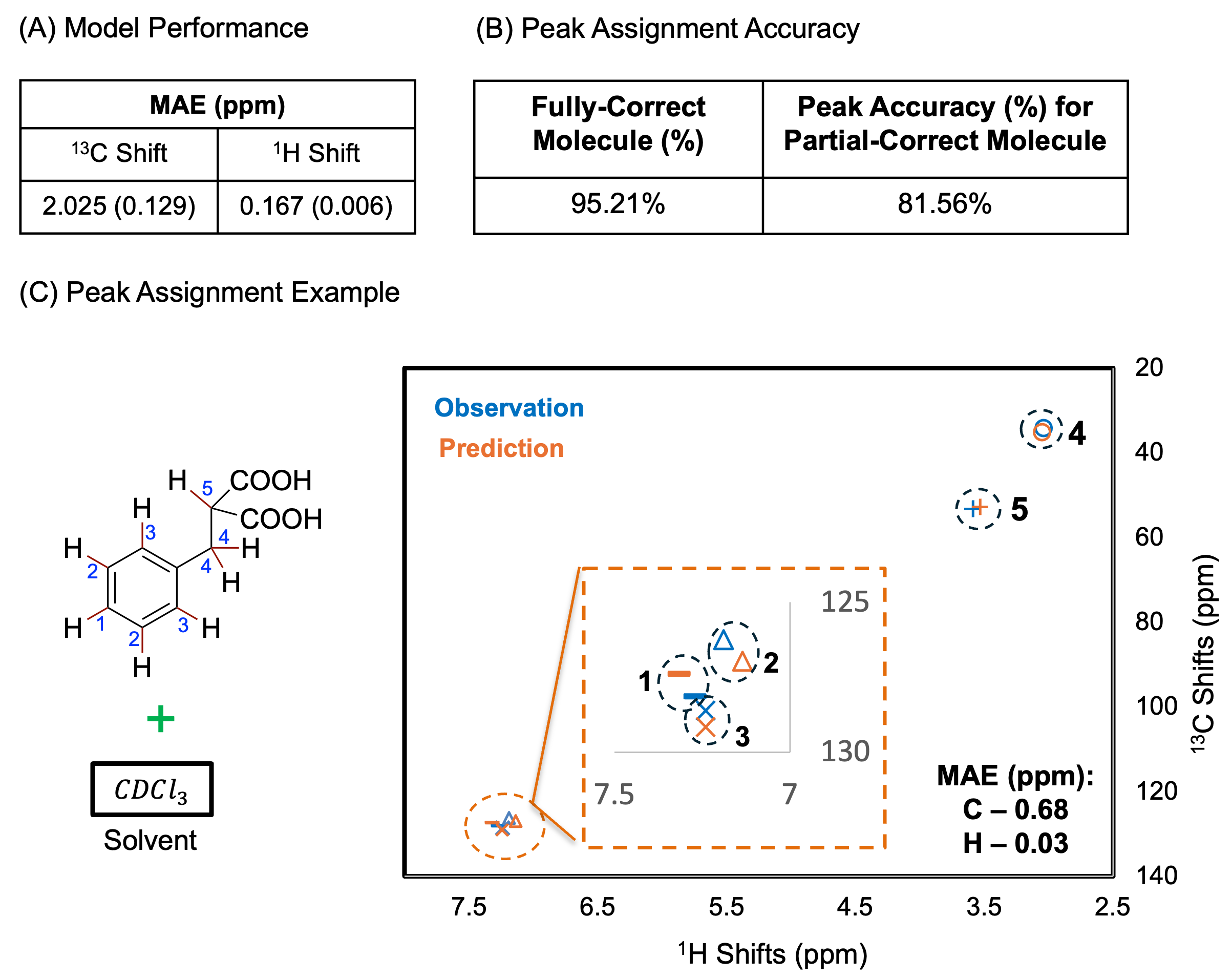}
    \caption{(A) MAEs of C--H shift prediction on test dataset. (B) Peak assignment accuracy by comparing algorithm-generated annotations with expert annotations. Out of the 479 molecules in the test set, 456 molecules have all peaks annotated correctly. For the remaining 23 molecules, 81.56\% of the peaks agree with expert annotations. (C) An example of using our model to accurately predict cross peaks and align them with experimental signals. The molecule is shown at the top-left, where each C--H bond is labeled with a numerical identifier. Notably, the symmetric pairs of bonds (labeled as ``2'', ``3'', and ``4'') are each expected to generate a single HSQC cross peak due to their structural equivalence. The HSQC cross-peaks predicted by our model (in orange) and their alignments to the experimental observations (in blue) are plotted in the right. The alignments are indicated by the dash circles.}
    \label{fig:result_summary}
\end{figure}

Figure \ref{fig:result_summary} summarizes the performance of our model on the tasks of HSQC cross-peak prediction and peak assignment, using an expert-annotated test dataset consisting of 500 molecules, with an average molecular weight of 398.98 Da., and an average number of 56.32 atoms. The annotation process involved three experienced experts with extensive knowledge in organic chemistry. For each molecule, two experts independently linked the observed cross-peaks from experiments to C--H bonds. If they agreed, the annotation was finalized. In cases of disagreement, the third expert reviewed and validated the annotations. Samples with poor quality, such as those with insufficient experimental resolution, were excluded from the test dataset for model evaluation, resulting in 479 high-quality annotations. For chemical shift prediction, our model achieved an MAEs of 2.05 ppm for $^{13}\text{C}$ shifts and 0.16 ppm for $^{1}\text{H}$ shifts. In terms of annotation accuracy, our model accurately annotated all peaks in 456 out of 479 molecules (95.21\%). For those 23 molecules that our algorithmic annotations do not fully agree with the experts, 81.56\% of the peak annotations still align. An example of the model outcome of peak prediction and assignment with correct solvent input is included in Figure \ref{fig:result_summary}.

\begin{figure}[!h]
    \centering
    \includegraphics[width=0.99\linewidth]{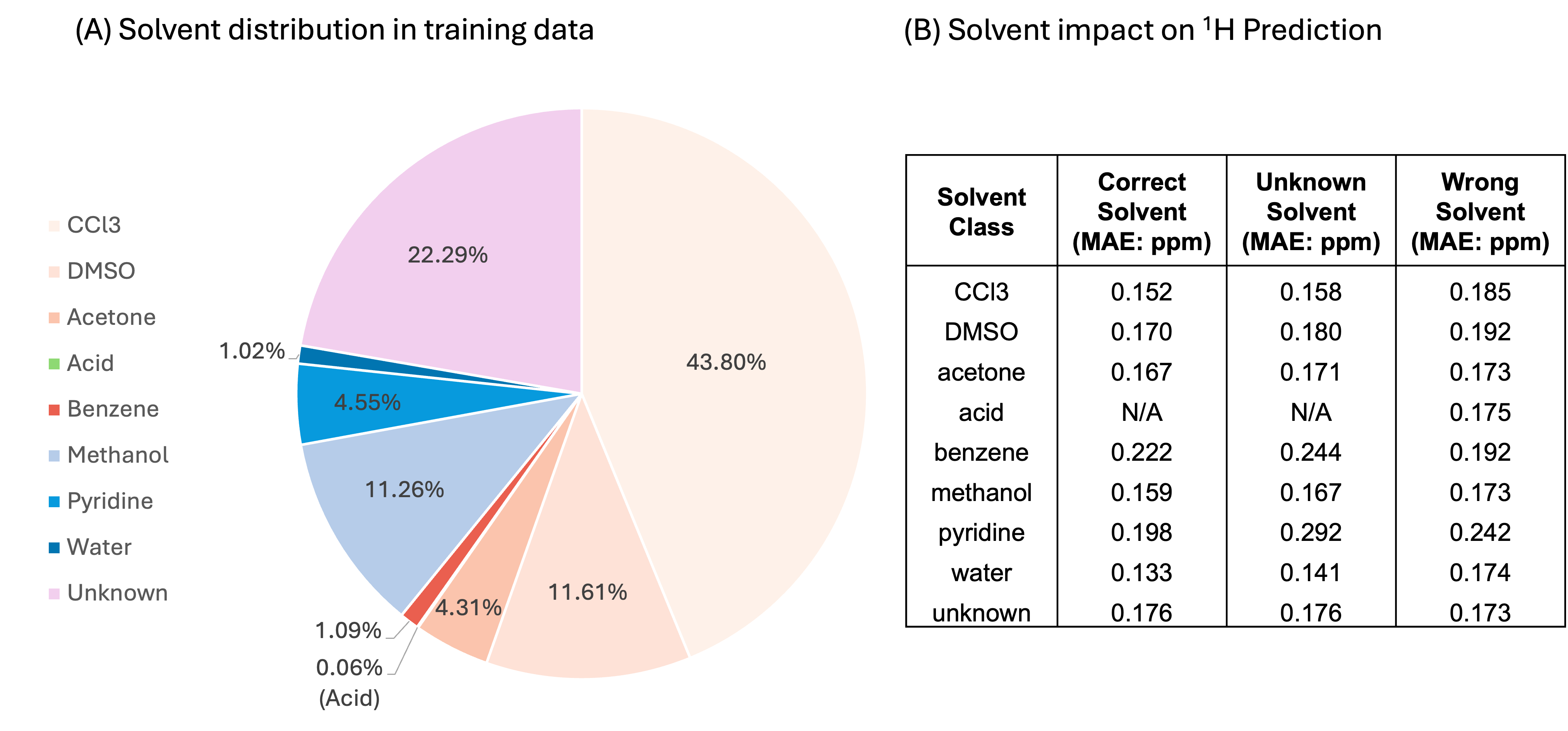}
    \caption{(A) The distribution of 9 solvent classes in the training dataset. (B) Solvent effect on proton shift prediction. When using the correct solvent information, the model provides the most accurate shift prediction. In most cases, specifying the solvent as ``unknown'' yields better performance, than using a wrong solvent as input. The acid solvent environment is marked as ``N/A'' in the table because it was not captured in the test dataset due to its low presence in the dataset. }
    \label{fig:solvent}
\end{figure}

To evaluate our model's ability to capture solvent effects, we tested different solvent encoders for carbon and hydrogen atoms. Empirical results indicate that a solvent encoder with a dimension of 32 is most effective at capturing proton shifts, while adding additional embeddings for carbon does not yield significant benefits. This partially aligns with expert expectations, as protons are known to be more sensitive to their solvent environment due to their high exposure to the surrounding molecular environment. By contrast, carbon atoms are less directly influenced by solvent interactions, as they are more deeply embedded within the molecular structure and shielded by surrounding electron clouds. Additionally, carbon atoms are not directly involved in hydrogen bonding, and their larger mass and lower sensitivity to external magnetic fields make their shifts less responsive to subtle solvent changes. Furthermore, since the 1D NMR data used for pre-training does not include solvent information, and some solvent groups (e.g., benzene, acid, etc.) in the HSQC data are underrepresented, as shown in Figure \ref{fig:solvent}(A), the influence of solvent on carbon shifts may be too subtle to capture at this stage. This remains as our ongoing area of investigation for further insights. To assess the effect of solvents on proton shifts, we report model performance using the true experimental solvent, a random solvent condition, and the "unknown" solvent condition, with the comparison shown in Figure \ref{fig:solvent}(B). The results demonstrate that using the correct solvent input yields the lowest prediction error, aligning most closely with experimental observations. This improvement in prediction is particularly significant for the CCl3, DMSO, and methanol solvent classes, likely due to their prominent presence in the dataset. The effect of water, on the other hand, despite constituting only 1.02\% of the data, can be effectively captured by the solvent encoder. This could be explained by water's similarity to methanol in its behavior as a hydrogen-bond donor or acceptor. Both solvents can strongly deshield protons in solutes, causing their NMR peaks to shift higher. The predictability of these hydrogen-bonding interactions may enable the model to generalize well, even with relatively few training examples for water. These findings underscore the promise of incorporating solvent environments into peak prediction models and highlight the need for more high-quality data with solvent information.


\subsection{Comparison with traditional tools}
\label{sc:trad_performance}
\begin{figure*}[!ht]
    \centering
    \includegraphics[width=1\linewidth]{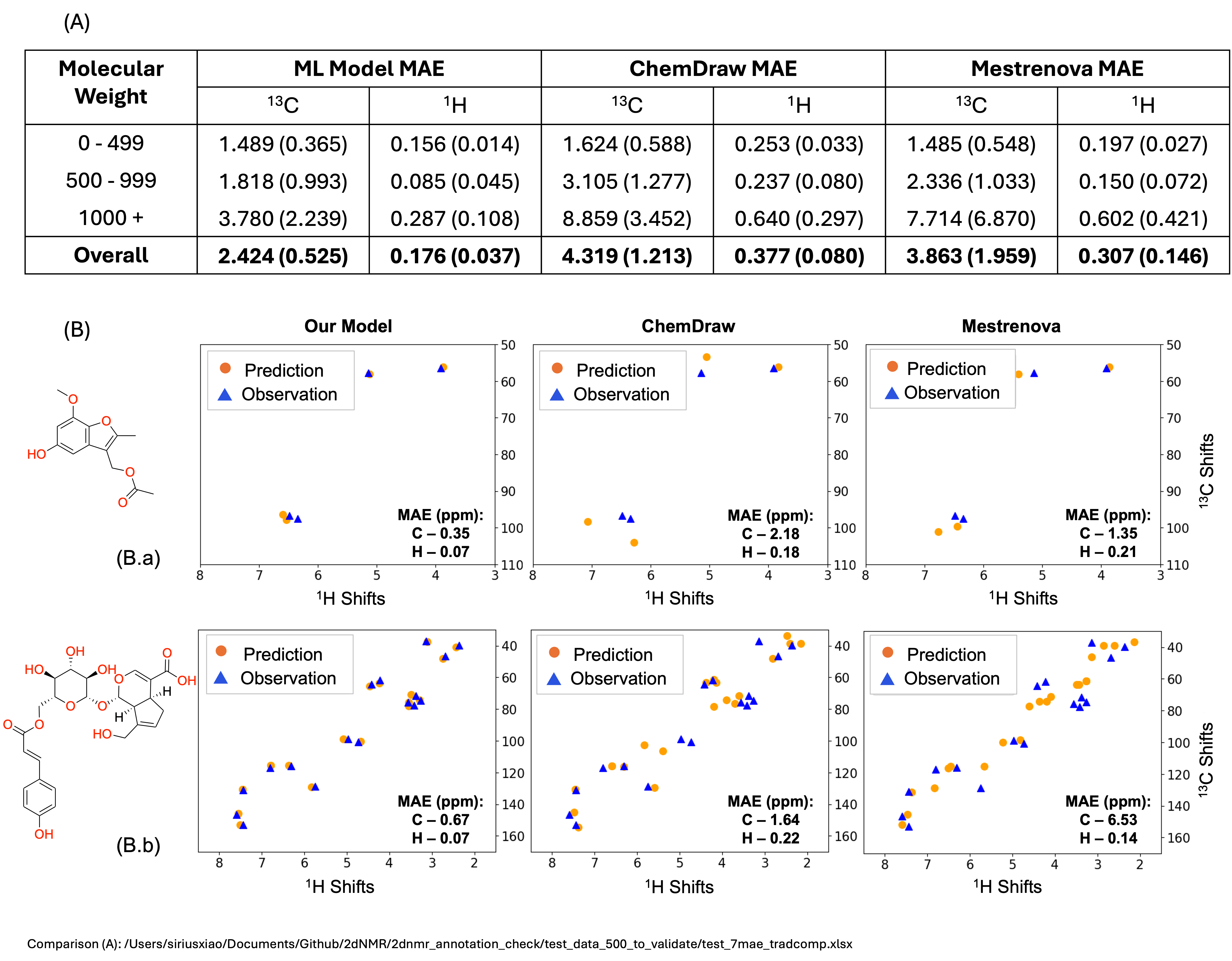}
    \caption{(A) Performance comparison between our proposed model and established traditional tools on randomly sampled molecules from the test dataset. Our model performs better across all molecular weight categories. The advantage of our approach is increasingly evident as molecular size increases. The overall result uses equal weight for the molecular weight categories. (B) Comparing our model, ChemDraw, and Mestrenova on two typical examples. A small molecule (a) with weight of $\sim$250 Dalton and a larger molecule (b) with weight of $\sim$500 Dalton. The observed experimental signals and the predicted signals are colored in blue and orange, respectively. The prediction error (MAEs) is shown in the bottom right corner of each plot. Our model performs better than ChemDraw and Mestrenova, and particularly excels in handling large molecules with complex conformations.}
    \label{fig:comp}
\end{figure*}

In organic chemistry, simulating HSQC spectra is crucial for analyzing experimental HSQC spectra, as it assists researchers in assigning the observed cross peaks to the C--H bonds in target molecules. Traditional approaches, including software solutions such as \textit{ChemDraw}\cite{mills2006chemdraw}, and \textit{Mestrenova}\cite{willcott2009mestre}, have long served as the primary resources for this task. We compared our model with \textit{ChemDraw} and \textit{Mestrenova} and the results are shown in Figure \ref{fig:comp}, which clearly demonstrate the superiority of our model.  \yl{Since these traditional methods only provide 1D NMR predictions, the cross peaks were identified and assigned by manually correlating the predictions from 1D 13C NMR and 1H NMR spectra, using atom connectivity as the basis for establishing the correspondence.
} We also provided two examples with different molecular sizes to visualize the comparison. 


\subsection{Performance by segmentation}
\label{sc:mw_sugar_performance}

To comprehensively evaluate the robustness and generalizability of our model, we conducted a segmentation analysis in the test dataset, assessing its performance across various molecular subcategories. Segmentation is crucial in this context because it allows us to determine whether the model performs consistently well across diverse molecular characteristics, rather than excelling in only a specific subset. We selected categories based on molecular weight and the presence of saccharides. These categories were chosen because NMR prediction gets increasingly challenging as the molecule gets larger, while saccharides represent a distinct chemical group with unique structural features. By examining the model's performance within these defined segments, we aim to demonstrate its universal applicability and robustness across a wide range of molecular types.

Molecular Weight (MW) is a general indicator of a molecule's complexity, encompassing varied geometries, bonding patterns, and the presence of isomers. These factors contribute to increased intramolecular interactions, resulting in spectral complexity such as closely spaced peaks or overlapping signals. Additionally, the increased number of spin-spin interactions within larger molecules necessitates more advanced NMR techniques to achieve sufficient resolution.\cite{foster2007solution} Furthermore, solubility issues can lead to weak signals, further complicating spectral analysis. Consequently, interpreting HSQC spectra for medium and large molecules is challenging. Therefore, there is a pressing need for a model that can effectively predict and analyze HSQC spectra for these complex molecules.

\begin{figure}[!h]
    \centering
    \includegraphics[width=0.90\linewidth]{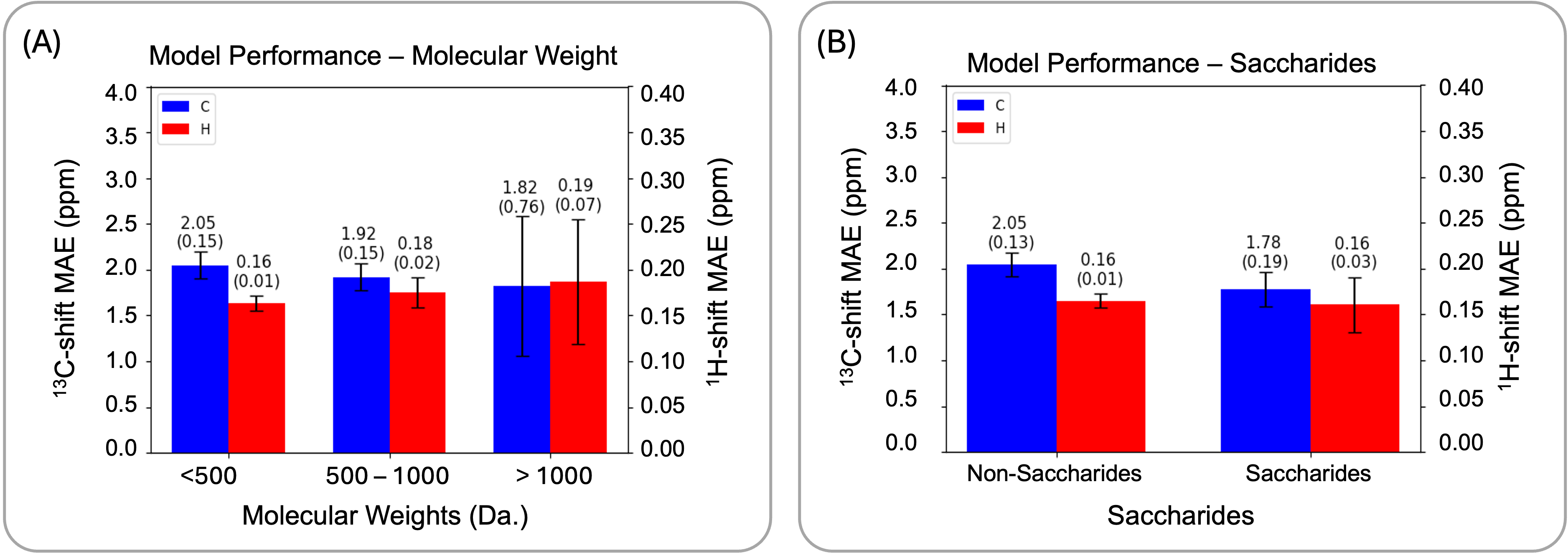}
    \caption{Model performance comparison on different segmented categories, including (A) molecular weights and (B) saccharides.}
    \label{fig:sugar_wt}
\end{figure}

Figure \ref{fig:sugar_wt}(A) showcases the stratified performance on this category, where the test molecules are grouped into three categories: small (MW < 500 daltons), medium (500 <= MW < 1000 daltons), and large (1000 daltons <= MW). On the task of predicting $^{1}\text{H}$ shifts of HSQC cross peaks, the model performs comparably across all groups, achieving excellent MAE of 0.16 - 0.19 ppm. On the task of predicting $^{13}\text{C}$ shifts, the model achieves an MAE of 1.93 ppm for medium-sized molecules. Our model demonstrates a good generalization power on large molecules and achieves an MAE of 2 ppm on predicting $^{13}\text{C}$ shifts for this category, despite the training data containing only a small proportion of large molecules ($\sim$2\%). 

Saccharides, or carbohydrates, play critical roles in various biological processes involved in energy source and storage, cell signaling, cell adhesion, cell recognition, structural integrity of cells and tissues, as well as cognitive functions and metabolic regulation.\cite{guillen2010carbohydrate,yu2018biological,dashty2013quick, fadaka2017biology} 
Despite their importance, elucidating the structures of saccharides is challenging due to their inherent structural complexity and diversity. This complexity arises from the diverse arrangements of monosaccharide units, varied anomeric configurations, and variable glycosidic linkages. Additionally, carbohydrates often lack the crystallinity required for high-resolution X-ray diffraction, unlike the well-defined crystalline structures of small molecules or proteins. Consequently, NMR spectroscopy, particularly through techniques such as HSQC, has emerged as an indispensable tool in unraveling the detailed structures of carbohydrates.\cite{brown2018solution, fels2022application} Forecasting HSQC cross peaks and aligning them with experimental data can assist in comprehending saccharide connectivity and stereochemistry, thus aiding in structural determination.

\begin{figure*}[!h]
    \centering
    \includegraphics[width=1\linewidth]{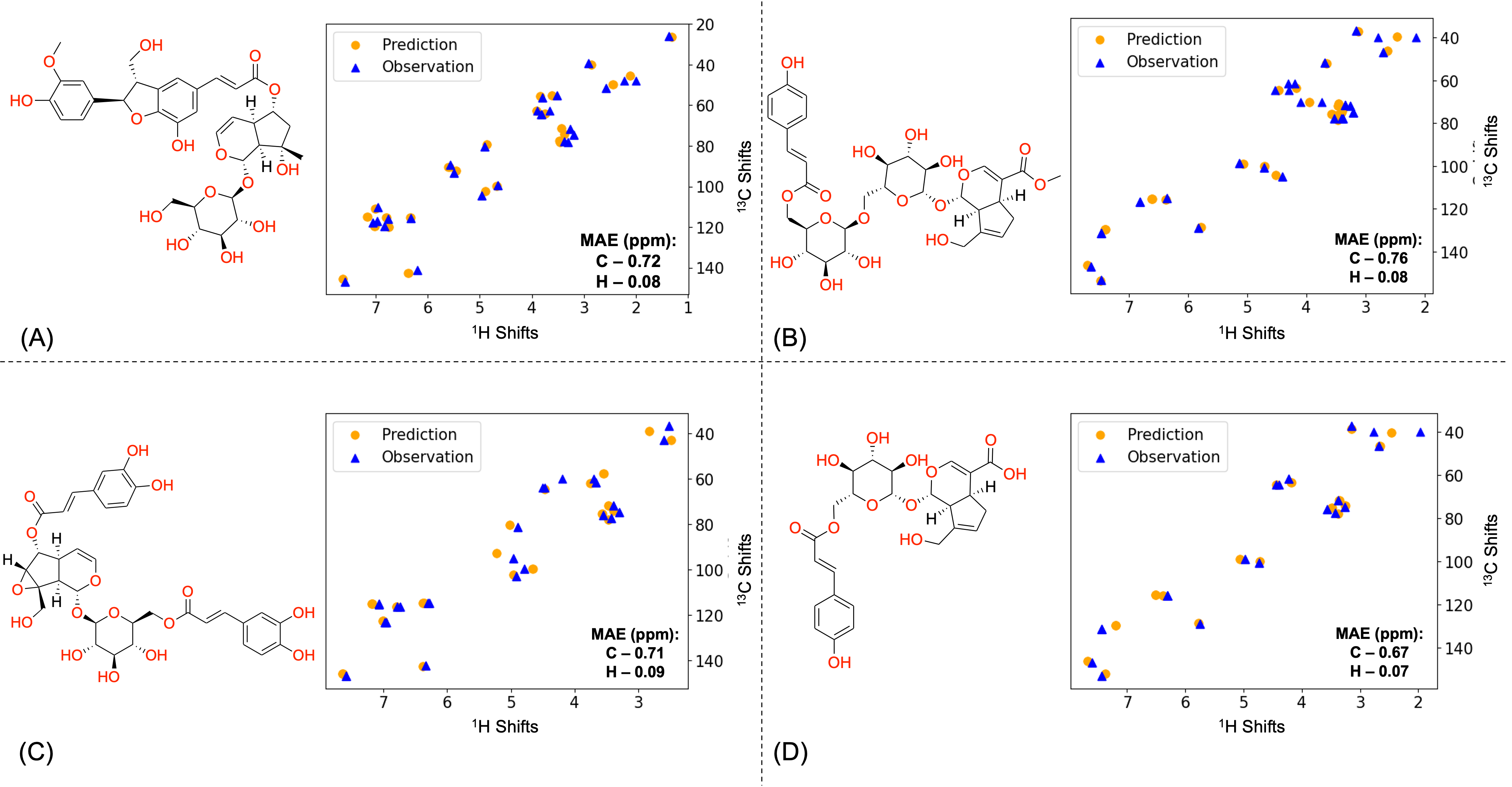}
    \caption{Exemplary demonstration of our model's performance on saccharides. The observed experimental signals and the predicted signals are colored in blue and
orange, respectively. The prediction error (MAEs) is shown in the bottom right corner within each plot.}
    \label{fig:sugar}
\end{figure*}

Our model demonstrates excellent performance in predicting HSQC cross peaks for saccharides molecules, yielding MAEs of 1.79 for $^{13}\text{C}$ shifts and 0.16 for $^{1}\text{H}$ shifts (see Figure \ref{fig:sugar_wt}(B)). 
This level of accuracy is consistent with the overall model performance, which demonstrates the model's robustness in handling complex saccharide structures.
Figure \ref{fig:sugar} shows the performance of our model on a few exemplar saccharides. These saccharides feature multiple ring structures and numerous stereogenic centers, contributing to the intricate nature of their HSQC spectra. Despite these inherent complexities, our model exhibits exceptional accuracy in predicting the HSQC cross peaks for these molecules. This robust performance underscores our model's capacity to navigate the complexities associated with saccharides, thereby emphasizing its versatility and effectiveness across various applications in the field.



\subsection{Effects of pre-training and fine-tuning}
\label{sc:mtt_iul_performance}

After pre-trained via MTT on the 1D NMR dataset, the model achieved the validation performance with MAEs of 0.210 ppm for $^{1}\text{H}$ NMR prediction and 2.228 ppm for $^{13}\text{C}$ NMR prediction. This success can be attributed to MTT which allows the model to effectively learn atomic latent features as well as local structural information by simultaneously performing $^{1}\text{H}$ and $^{13}\text{C}$ NMR shift predictions. This helps us surpass the problem with limited annotated HSQC data. 
However, when directly deploying the pre-trained model on HSQC test dataset, the model MAEs increase to 1.397 ppm and 2.822 ppm for $^{1}\text{H}$ and $^{13}\text{C}$ shifts, respectively. These relatively large MAEs are expected as the data distribution of the HSQC dataset (76.34\% small molecules and 90.33\% non-saccharides) differs significantly from that of the 1D NMR dataset (98.80\% small molecules and 99.95\% non-saccharides). In addition, the HSQC cross peaks involve interactions beyond simple pairings of 1D $^{13}\text{C}$ and $^{1}\text{H}$ shifts, requiring a deeper understanding of interactions between atoms. Finally, the frequent absence of solvent labels in the 1D NMR dataset prevents the model from learning solvent effects.

Nevertheless, the pre-training via MTT offers a robust foundation for fine-tuning the model via unsupervised transfer learning. With each iteration, we observed a reduction in model errors. The performance improvement is more pronounced during the initial iterations and gradually diminishes. By the fifth iteration, the improvement became marginal, indicating the convergence of fine-tuning. Finally, the fine-tuned model achieves MAEs of 0.165 ppm and 2.05 ppm for $^{1}\text{H}$ and $^{13}\text{C}$ shifts, respectively. Throughout the transfer learning process, the model was trained to gain a more profound understanding of solvent effects and complex C--H interactions due to intricate molecular structures.




\section{Discussion}

In this study, we introduce a novel framework to develop machine learning techniques for predicting C--H cross peaks in HSQC spectra, The framework enables us to tackle two major challenges in this avenue. The first challenge is the scarcity of annotated HSQC data for training machine learning models. The second challenge is that collecting large volumes of annotated HSQC data is labor-intensive and requires highly trained personnel. In implementing our framework, we developed a model combining a GNN with a solvent encoder. The GNN is trained to generate atomic embeddings that encapsulate both the local and global chemical environments of each atom, which is crucial for accurate chemical shift predictions. The atomic embeddings are combined with the solvent embedding produced by the solvent encode, which allows our model to learn the influence of solvent on chemical shifts. The combined embeddings are mapped by \yl{the Multi-Layer Perceptron (MLP)} modules to HSQC chemical shifts. Our framework employs a two-stage transductive strategy to train the model while addressing the aforementioned challenges. In the first stage, we use a large amount of annotated 1D NMR data to pre-train the model via Multi-Task learning. This enables the model to adeptly grasp the intricate relationship between atomic interactions and NMR signals, laying a robust foundation for the subsequent stage. Next, the model is refined on a set of unlabelled HSQC spectra via Iterative Unsupervised Learning, enhancing the model's capability in predicting and interpreting HSQC spectra. Our final model achieves MAEs of 0.165 ppm and 2.05 ppm for $^{1}\text{H}$ and $^{13}\text{C}$ shifts respectively, while accurately assigning cross peaks. It demonstrates a consistent performance across various molecular weight and saccharide categories, significantly outperforming the traditional methods, and shows convincing generalization capabilities to less represented samples from the training dataset. In the future, we plan to refine our model by developing 3D-GNN models that are able to consider 3D structural information such as spatial orientation and conformational flexibility. This enhancement should enable us to handle other 2D NMR spectra, such as Correlation Spectroscopy and Nuclear Overhauser Effect Spectroscopy, thus broadening its applicability and making a more substantial contribution to the field of chemical analysis.

\section{Methods}

In this section, we explain the components of our model and the training strategy in detail.

\subsection{Data}
The pre-training dataset used in the MTT process is a 1D NMR dataset from NMRShiftDB2\cite{steinbeck2003nmrshiftdb}, which contains $\sim$24,000 annotated NMR spectra collected from 22,663 distinct molecules. The datasets used in the unsupervised transfer learning process consist of a training dataset containing $\sim$19,000 experimental HSQC spectra and a validation dataset containing $\sim$5,000 HSQC spectra, collected from HMDB \cite{wishart2022hmdb} and CH-NMR-NP \cite{j-resonance}. \yl{All data was accessed in September 2023, with no evidence of potential bias. To prevent data leakage in the validation dataset, duplicated spectra and molecules were removed. RDKit package in Python is used to perform sanity check for all SMILES strings to generate valid molecular topology graph.} To quantitatively evaluate our model, we built a test dataset by randomly selecting 500 spectra and manually annotating them to establish the ground truth (see Section 2.1 for the annotation process). Additionally, to compare our model with two conventional tools (ChenDraw and Mestrenova) in chemistry, we randomly selected several molecules from this test dataset, consisting $\sim$150 cross-peaks. Since it is labor-intensive to derive HSQC shifts from molecular formulas using these established tools, stratified sampling was used to select these samples, ensuring the coverage of different molecular weight groups (0-499 Dalton, 500-999 Dalton, and 1000+ Dalton). The comparison results are presented in Section \ref{sc:trad_performance}. 





\subsection{2D NMR prediction model}
\label{sec:framework}

As illustrated in Figure \ref{fig:framework_training_strategy}(A), our model contains a GNN component for encoding molecular features and a solvent encoder component for embedding solvent information. The GNN component learns atomic embeddings that capture both the local and global chemical environments of each atom, which are essential for understanding the observed NMR chemical shifts. The learnt atom representations are expanded by the solvent embedding, and then are mapped to $^{13}\text{C}$ and $^{1}\text{H}$ cross peaks by a \yl{MLP} component. 


\subsubsection{GNN}
\begin{figure*}[!h]
\centering    
\includegraphics[width=0.98\linewidth]{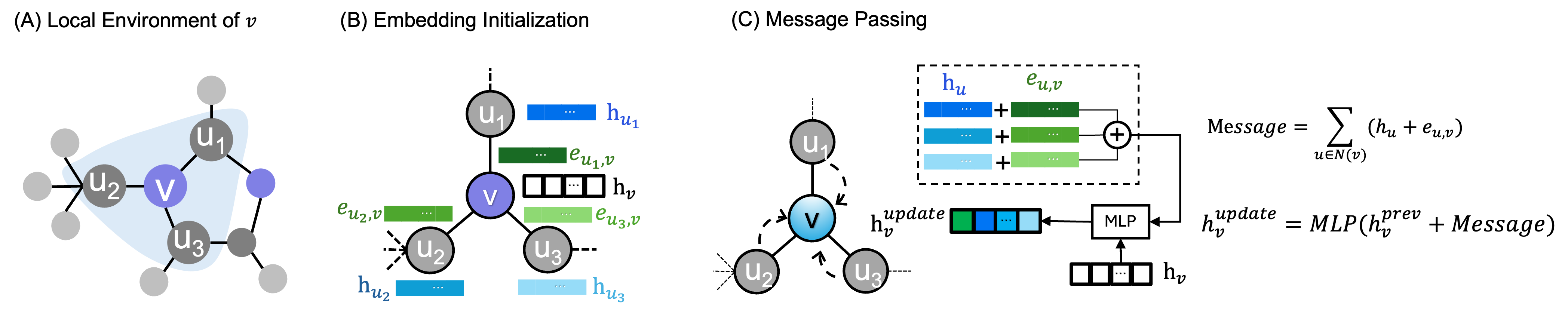}
\caption{Illustration of message passing and node representation updates in a GNN layer. (A) For a given center atom in a molecular graph, the local environment contains the neighborhood atom that is directly bonded to the center atom. (B) For a center node ($v$), initial random representations are assigned to this node, its neighboring nodes ($u_1$, $u_2$, $u_3$), and their connecting bonds ($e_{u_1, v}$, $e_{u_2, v}$, $e_{u_3, v}$). (C) Message Passing and node update. Representations of all neighboring nodes and edges are aggregated and integrated to form a message to the center node. The representation of the center node is then updated to incorporate this message and information from its previous state. }
\label{fig:gnn_mp}
\end{figure*}

A molecule can be represented by a graph $G = (V, E)$, where $V$ is the node set representing atoms and $E$ is the edge set representing chemical bonds. Three features are provided for each node: atomic type, chirality, and hybridization. Also, two features are considered for each edge: bond type and bond direction. Bond types include Single, Double, Triple, and Aromatic, each reflecting a distinct configuration of electron sharing between atoms. Bond direction includes None, EndUpRight, and EndDownRight, primarily representing stereochemistry in double bonds. Each atom's feature vector is embedded into a representation vector by a learnable encoder. Similarly, each edge's feature vector is embedded into a representation vector of the same length by another learnable encoder.
Then, a GNN model \cite{scarselli2008graph, micheli2009neural, gilmer2017neural,  zhou2020graph, reiser2022graph} utilizes the message passing mechanism to iteratively refine the representation of each node based on information from its neighbors and connected edges. This mechanism allows the learnt node representation to effectively capture structural context, reflecting the foundational principles of atomic interactions. 
Our implementation of the message passing mechanism is illustrated in \ref{fig:gnn_mp}. It iterates for a predefined number of layers $L$, facilitating the propagation of information throughout the graph. Consequently, each node can gradually accumulate information from a wider neighborhood across successive layers. This allow the final representation of each node to capture both local and global structural information. Our model features 5 GNN layers, with an atomic embedding dimension of 512.




\subsubsection{Solvent encoder}
Since the solvent has a profound impact on NMR chemical shifts, we incorporated a trainable solvent encoder component into our model to accurately capture this influence. We identified the following 9 principal solvent groups based on their prevalence in our dataset and domain-specific understandings of their distinct impacts on NMR shifts.  These groups include trichloromethane, dimethyl sulfoxide, acetone, acids, benzene, methanol, pyridine, water, and an additional category to encompass any unspecified solvents from our dataset (termed "unknown"). The solvent encoder transforms each discrete solvent group $i$ into a unique, dense feature vector $S_i^d$, where $d$ is the embedding dimension. These learnable vectors are optimized alongside other model parameters during training, resulting in representations that accurately reflect the impact of each solvent class. Given the different sensitivities of carbon (C) and hydrogen (H) nuclei to solvent environments, different embedding dimensions $d$ can be chosen to tailor the solvent effect modeling for each nuclei type. A larger embedding dimension $d$ allows the embedding to more effectively capturing the nuanced influence of solvents on NMR shifts. In our implementation, the solvent embedding dimension for hydrogen (H) is set to 32.

\subsubsection{Atomic NMR shift prediction}

Finally, the embedding of each atom $h_v^{(L)}$ and the solvent embedding $S_i^d$ for each solvent class $i$ are concatenated to produce a holistic representation of the atom within the context of its molecular structure and the given solvent. This combined representation is subsequently processed by a MLP network to predict the NMR shifts for the atom:

\begin{equation}
y_v = \text{MLP}(h_v^{(L)} \oplus S_i^d)
\end{equation}

where $y_v$ is the predicted chemical shift of atom $v$, $h_v^{(L)}$ is the atom level embedding produced by GNN, $S^d$ is the solvent embedding, and $\oplus$ is the concatenation operation. By integrating solvent embedding and atomic embedding, the model effectively combines intrinsic molecular properties and solvent effects, enhancing its ability to predict atomic NMR shifts accurately. 

Two separate MLP modules are used for predicting $^{13}\text{C}$ and $^{1}\text{H}$ shift in the cross peak predictions, respectively. Each C atom can bond up to 4 H atoms. When bonded to one, three, or four H atoms, a C atom typically shows only one cross peak in an experimental spectrum. However, when a C atom is connected to two H atoms, up to two cross peaks may be observed, depending on the chiral center. Consequently, a C atom can exhibit at most two $^{13}\text{C}$ and $^{1}\text{H}$ cross peaks. In light of this observation, one MLP module is dedicated to predicting the $^{13}\text{C}$ shifts and another MLP module for the corresponding  $^{1}\text{H}$ shifts. For cross peak predictions, the $^{13}\text{C}$ shifts are predicted using the embeddings of C atoms. The corresponding $^{1}\text{H}$ shift predictions for each C atom incorporate aggregates of embeddings from all bonded H atoms, resulting in two predictions that are typically very similar when only one cross peak is theoretically possible. This design enhances the model's accuracy in predicting $^{1}\text{H}$ shifts by leveraging the C atom-centered aggregation of the H atom context. By integrating the contextual dynamics around each C atom, the model provides a more detailed and accurate mapping of hydrogen environments, crucial for pinpointing precise cross peaks in complex HSQC spectra. In our implementation, we used 2 MLP layers, with the hidden dimensions to be 128 and 64 respectively.


\subsection{Training strategy}
\label{sec:training}

The cross peaks are notably sparse in an HSQC spectrum, where typical resolutions for $^{13}\text{C}$ and $^{1}\text{H}$ shifts are 0.1 and 0.01 ppm, respectively. A typical HSQC spectrum can include ~20,000 readings, covering $^{13}\text{C}$ shifts from 0 to 200 ppm and $^{1}\text{H}$ shifts from 0 to 10 ppm. However, almost all of these readings are zeros, with only a small fraction representing the potential cross peaks of C--H bonds, crucial for molecular structure analysis.  Moreover, the scarcity of annotated HSQC data, particularly the labor-intensive annotations that link cross peaks to C--H bonds, makes model training difficult. To deal with this issue, we deployed MTT to pre-train the model using an extensive annotated 1D NMR dataset (Fig. \ref{fig:framework_training_strategy}B). This step acclimates the model with a broad range of molecular structures and their chemical shifts, and enables it to capture the intricate interplay between molecular structures and their NMR characteristics. Subsequently, we utilize an unsupervised strategy to refine the model iteratively on the HSQC dataset (Fig. \ref{fig:framework_training_strategy}C). Through iterative cycles of prediction, annotation, and re-training, the model progressively enhances its understanding of the complex relationships and patterns within the HSQC spectra, thus improving its predictive accuracy and providing precise cross peak alignments. By combining the MTT and unsupervised transfer learning, we extend our annotation capabilities from 1D to 2D data, thereby enhancing the model's predictive power and utility as a robust tool for NMR spectra analysis.

\subsubsection{Pre-training on 1D NMR data}
\label{sec:multitasking}

In the pre-training phase, we utilized approximately 24,000 annotated 1D NMR data points. Among these, around 22,000 samples exclusively feature $^{13}\text{C}$ shifts, approximately 400 samples solely exhibit $^{1}\text{H}$ shifts, while roughly 1,600 samples contain both $^{1}\text{H}$ and $^{13}\text{C}$ shifts.  
To train the model effectively for predicting both $^{1}\text{H}$ and $^{13}\text{C}$ shifts, we adapt the MTT approach, which enables simultaneous training on multiple related tasks. When the input data contains $^{13}\text{C}$ shifts, the model predicts only carbon shifts and assesses the errors between the predicted and actual values. Conversely, when the data sample contains $^{1}\text{H}$ shifts, the hydrogen shift prediction module is activated. In both scenarios, the embeddings of $^{13}\text{C}$ and $^{1}\text{H}$ atoms in the GNN module are updated simultaneously, benefiting from the message passing mechanism. Therefore, the learnt representations implicitly contain a basic understanding of C--H relationships, essential for the interpretation of HSQC data. However, the relative scarcity of $^{1}\text{H}$ shift data, due to the difficulties in accurately obtaining and extracting peaks $^{1}\text{H}$ from experimental data, complicates the training process as focusing extensively on one type of shift could compromise the model’s ability to accurately predict the other. To handle this problem in the MTT training, we performed over-sampling on a subset of data that contain both $^{1}\text{H}$ and $^{13}\text{C}$ shifts, and those containing only $^{1}\text{H}$ shifts. Consequently, the learned representations develop a fundamental understanding of C--H relationships, crucial for interpreting HSQC data effectively. This integration of learned atomic relationships streamlines the transition to HSQC cross peak predictions, thereby enhancing the model's accuracy and efficiency in analyzing HSQC spectra.

\subsubsection{Unsupervised fine-tuning on HSQC data} 
\label{sec:autoregressive}

The model pre-trained on the 1D NMR dataset has limited ability to predict HSQC cross-peaks from molecular structures due to the differences in data pre-processing and data distribution. First, in the 1D NMR data from NMRShiftDB2, the chemical shifts of non-singlet peaks are averaged as ground truth. For example, whether the group is methine (-CH), methylene (-CH2), or methyl (-CH3), the proton shifts may be averaged into a single value. In contrast, HSQC data captures C-H bonds and typically displays two cross-peaks for prochiral methylene groups due to the different environments of the two hydrogens. Additionally, the proton chemical shifts of HSQC or HMQC cross peaks represent $^{13}\text{C}$-bound protons, whereas the signals in the $^{1}\text{H}$ mainly represent $^{12}\text{C}$-bound protons, potentially leading to subtle differences in chemical shift values\cite{schroeder9}. Second, the molecule distributions in our 1D NMR data and HSQC data exhibit significant differences. The HSQC dataset comprises 76.34\% small molecules and 90.33\% non-saccharides, whereas the 1D NMR dataset contains 98.80\% small molecules and 99.95\% non-saccharides. Lastly, solvent information is not available in 1D NMR dataset, and is recorded as ``unknown'' in the modeling framework, whereas most molecules in the HSQC dataset are associated with known solvent environments. This makes the fine-tuning step essential for the success of our solvent-aware framework.
However, the HSQC dataset is not annotated. In response, we implement an unsupervised training strategy (Fig. \ref{fig:framework_training_strategy}C), which iterates between (a) aligning cross peak prediction from the model with the experiment observations to annotate the HSQC data and (b) using the newly acquired annotations to fine-tune the NMR prediction model, until convergence.

\subsubsection{Pseudo-annotation of HSQC}
At the end of each round in the unsupervised learning process, the model's predicted signals are aligned with the experimental observations to create pseudo-labels. In straightforward cases where the number of C--H bonds in a molecular graph matches the observed HSQC cross peaks, the Hungarian algorithm \cite{kuhn1955hungarian, munkres1957algorithms} is used. This classic optimization technique solves assignment problems by minimizing the cost of matching a set of predictions to a set of observations. In the context of NMR analysis, the``cost'' is defined as the discrepancy between the predicted chemical shifts and the actual shifts observed experimentally. By systematically reducing these differences, the Hungarian algorithm achieves an optimal one-to-one correspondence between predicted shift pairs and experimental signals, even in complex scenarios with potential signal overlap.

However, in most cases, the number of C--H bonds within a molecule exceeds the number of signals recorded, making peak alignment more difficult. This mismatch in numbers arises from several factors: firstly, rotational equivalence can reduce the number of signals, with a single peak representing all three C--H bonds for methyl groups; secondly, symmetrical molecular structures can result in a single detectable signal for multiple symmetric C--H bonds, as seen in benzene molecule where only one peak represents all six C--H bonds; lastly, in highly complex molecules, overlapping signals obscure some peaks, reducing the detectability of individual C--H bonds from experiments.

To overcome this issue, we utilize the graduated assignment algorithm \cite{gold1996graduated, wang2023motif}, which facilitates matching between graphs of different node counts, making it particularly suitable for this scenario. In this algorithm, our model's predicted C--H shifts $(C^i, H^i)^N _{i=0}$ and the observed C--H signals $(C^j, H^j)^M _{j=0}$ of each molecule are conceptualized as points on a 2D plane, where $N$ and $M$ are numbers of predicted and observed C--H shifts respectively. These points are then treated as vertices in two fully connected graphs, $G_1$ for predicted shifts and $G_2$ for observed signals. The similarity between nodes is defined as the inverse of differences between predicted chemical shifts (node in $G_1$) and observed shifts (node in $G_2$). Specifically, for each predicted shift, we compute its difference with every observed shift, where a smaller difference indicates a higher similarity. The derive the assignment matrix $A$ where each element $A_{uv} \in \{0, 1\}$ indicates whether node $u$ in $G_1$ matches with node $v$ in $G_2$, the algorithm first finds the soft matching matrix that relaxes the binary constraint $A_{uv} \in \{0, 1\}$ to a continuous range $[0, 1]$, then converts it into hard assignment in a greedy way, enabling one-to-many matching. 

\section*{Data Availability}
HSQC data used for model performance testing is available \hyperlink{https://drive.google.com/drive/folders/1wQxk7mnIwi5aAGaF34_hk7xo6IeEh-IE?usp=sharing}{here (via Google Drive)}, and the training and validation datasets will be available upon request. 

\section*{Code Availability}
The code is available in Github: \hyperlink{https://github.com/siriusxiao62/2dNMR.git}{https://github.com/siriusxiao62/2dNMR.git}

\section*{Conflicts of interest}
There are no conflicts of interest to declare. 

\section*{Acknowledgements}
This work was supported by GlycoMIP, a National Science Foundation (NSF) Materials Innovation Platform funded through Cooperative Agreement DMR-1933525, as well as NSF OAC 1920147.



\balance

\newpage
\bibliography{nature} 
\bibliographystyle{naturemag} 

\end{document}